\documentclass[conference]{IEEEtran}
\IEEEoverridecommandlockouts
\usepackage{cite}
\usepackage{amsmath,amssymb,amsfonts}
\usepackage{algorithmic}
\usepackage{graphicx}
\usepackage{textcomp}
\usepackage{xcolor}
\def\BibTeX{{\rm B\kern-.05em{\sc i\kern-.025em b}\kern-.08em
    T\kern-.1667em\lower.7ex\hbox{E}\kern-.125emX}}

\begin{document}

\title{Evolving spiking neuron cellular automata and networks to emulate in vitro neuronal activity
\thanks{This work was partially funded by the SOCRATES project (Norwegian Research Council, IKTPLUSS grant agreement 270961) and the DeepCA project (Norwegian Research Council, Young Research Talent grant agreement 286558.}
}

\author{\IEEEauthorblockN{J\o{}rgen Jensen Farner\IEEEauthorrefmark{1}, H\r{a}kon Weydahl\IEEEauthorrefmark{1}, \\
Ruben Jahren\IEEEauthorrefmark{1}}
\IEEEauthorblockA{\textit{Department of Computer Science} \\
\textit{Oslo Metropolitan University}\\
Oslo, Norway \\
\IEEEauthorrefmark{1}These authors contributed equally.} \\
\IEEEauthorblockN{Stefano Nichele\IEEEauthorrefmark{2}}
\IEEEauthorblockA{\textit{Department of Computer Science} \\
\textit{Oslo Metropolitan University}\\
Oslo, Norway \\
\textit{Department of Holistic Systems}\\
\textit{Simula Metropolitan}\\
Oslo, Norway \\
stenic@oslomet.no}
\hspace{7.31cm} \IEEEauthorrefmark{2}Co-senior authors
\and
\IEEEauthorblockN{Ola Huse Ramstad}
\IEEEauthorblockA{\textit{Department of Neuromedicine and Movement Science} \\
\textit{Norwegian University of Science and Technology}\\
Trondheim, Norway \\
ola.h.ramstad@ntnu.no}\\ [0.4cm]
\IEEEauthorblockN{Kristine Heiney\IEEEauthorrefmark{2}}
\IEEEauthorblockA{\textit{Department of Computer Science} \\
\textit{Norwegian University of Science and Technology}\\
Trondheim, Norway \\
\textit{Department of Computer Science} \\
\textit{Oslo Metropolitan University}\\
Oslo, Norway \\
kristine.heiney@oslomet.no}
}

\maketitle

\thispagestyle{plain}
\pagestyle{plain}

\begin{abstract}
Neuro-inspired models and systems have great potential for applications in unconventional computing.
Often, the mechanisms of biological neurons are modeled or mimicked in simulated or physical systems in an attempt to harness some of the computational power of the brain.
However, the biological mechanisms at play in neural systems are complicated and challenging to capture and engineer;
thus, it can be simpler to turn to a data-driven approach to transfer features of neural behavior to artificial substrates.
In the present study, we used an evolutionary algorithm (EA) to produce spiking neural systems that emulate the patterns of behavior of biological neurons in vitro.
The aim of this approach was to develop a method of producing models capable of exhibiting complex behavior that may be suitable for use as computational substrates.
Our models were able to produce a level of network-wide synchrony and showed a range of behaviors depending on the target data used for their evolution, which was from a range of neuronal culture densities and maturities.
The genomes of the top-performing models indicate the excitability and density of connections in the model play an important role in determining the complexity of the produced activity.
\end{abstract}


\section{Introduction}


Unconventional computing models and systems taking inspiration from biological systems have great potential for enabling the development of more efficient, flexible, and scalable computational systems.
However, biological systems are self-organizing and have optimized their behavior through natural evolution \cite{Walker2013}, making their mechanisms challenging to unravel and emulate in engineered systems.
The advent of microelectrode arrays (MEAs) to observe the electrophysiological activity of populations of neurons non-destructively over weeks or months \cite{Obien2015} enables the observation of such a self-organizing system.
Observing target systems in this way opens the door to taking a more top-down data-driven evolutionary approach to producing models emulating their behavior with no concern for the underlying mechanisms.
The ability to create an engineered system showing similar outward behavior as the target biological system may then produce good computational substrates, capturing the life-like behavior without the need to reproduce its precise origins \cite{heiney2020artificial,LTL2021}.

Unconventional computing frameworks, such as reservoir computing (RC) \cite{Jaeger2001,Maass2002}, aim to take advantage of the dynamic traits of a given substrate to perform computational tasks.
RC was originally developed as a method to avoid the difficult task of training recurrent neural networks but has since been expanded: by training a single linear readout layer, the nonlinear dynamics of any high-dimensional driven system, even a physical reservoir, can be tapped into to solve complex tasks \cite{Tanaka,Schrauwen2007,Horsman2014}.
It is generally understood that computational substrates used in such frameworks must be constructed within a certain dynamic regime to show the desired behavior allowing them to perform as ``good'' computational substrates; however, for many complex substrates, both theoretical and physical, designing the substrate within this regime poses a challenge.

In this study, we sidestep this design question by starting with a substrate we assume shows features beneficial for computation: biological neurons.
To develop theoretical computational substrates for use in RC and other similar computational frameworks, we emulate the spiking behavior of populations of biological neurons in vitro by evolving cellular automata (CAs) and network models of spiking neurons to match their spatiotemporal distribution of activity.
On the basis of the assumption that the activity recorded from these biological neural systems is representative of computationally beneficial mechanisms at play in producing this activity, it is reasonable to consider that other systems producing similar behavior by different mechanisms may also be computationally effective, and this type of approach may produce a better starting point for our computational models.

Various optimization approaches have been applied to the data-driven parameter tuning of neuron models to match recorded electrophysiological data, as reviewed by Van Geit et al.~\cite{VanGeit2008}.
In this type of approach, a tuning algorithm is applied to select parameters that yield the output spiking behavior observed in recordings of biological neurons, with certain constraints placed on the parameters to keep the models within the realm of physiological possibility.
The choice of optimization approach depends on the selected neuron model, 
as well as the form of the experimental data, which can be targeted for simple spike trains or down to details of spike shape (e.g., \cite{Lynch2015,Druckmann2007,Neymotin2017}).
(See also \cite{VanGeit2016} for a general Python package to perform such optimizations.)

However, in many of these prior studies on the data-driven optimization of neuron model parameters, the focus has been on reproducing the behavior of single neurons, often to match the spike train pattern or the response of the neuron to various electrical stimuli.
In contrast, little work has been done on the emulation of population-level dynamics using metaheuristic methods \cite{Herzog2007,Pandarinath2018}.
There are great challenges associated with scaling neuron models, as the space of behaviors is vast and the dependence on the system parameters is complex and often unpredictable, with many very different combinations of parameters yielding the same behavior.
In a set of well-known experiments, Prinz et al.~\cite{Prinz2004} showed that even in a relatively simple three-cell model of the pyloric network of crustacean stomatogastric ganglion, vastly different sets of system parameters could yield the same characteristic triphasic motor pattern observed experimentally.
From an evolutionary computation perspective, this means that roughly the same phenotype can be obtained with rather different genotypes.

This idea that similar behaviors can arise from disparate systems has interesting implications, both biologically and computationally.
As noted by Prinz et al.~\cite{Prinz2004}, their results suggest that individual biological neurons can ``self-tune'' to achieve a target excitability, and that similar tuning can occur at a population level.
This means that there is no single solution to achieve target network behavior; rather, networks of neurons can flexibly tune among these solutions depending on other existing constraints or changes in the system and continue to produce the needed activity.
From a computational perspective, this also means that there are many good ways for neural systems to perform computations, and if we wish to emulate the computational behavior of such systems, we perhaps need not replicate the exact parameters of each specific system we seek to emulate.
Instead, we can find more flexible ways to produce the same type of behavior as our target system, without concern for the physiological constraints that occur in biological systems.

In this study, we evolved simple networks and cellular automata (CAs) composed of spiking neurons to emulate features of spiking activity recorded from networks of biological neurons in vitro.
The biological data was obtained in an earlier study by Wagenaar et al.~\cite{Wagenaar2006repertoire} in an investigation of how the plating density of in vitro cultures affects their patterns of activity.
Our goal is to attain systems models that are able to produce behavior comparable to that of biological neurons, despite the substantially simpler mechanisms at play in our models---for example, the simpler threshold-based firing of the neurons, the limited size of the networks, the homogeneity of the components, and the lack of a synaptic model.

In this way, we aim to produce behavior akin to that of actual neurons but using much simpler mechanisms than those used in biological systems.
We expect that these models may serve as a good starting point for computational reservoirs---assuming the in vitro neuronal cultures exhibit computationally beneficial behaviors, our models may therefore capture, to a degree, the computational essence of the biological system we emulate.
This approach makes no a priori assumptions about the suitability of the resulting systems for a certain computational task \cite{Variengien2021} or their performance on benchmarking tasks \cite{Glover2021}, but future work will involve assessing our evolved systems from this type of computational perspective.

The next sections give an overview of the model design and the evolutionary process. Section \ref{secResults} then presents and discusses the results of our experiments, and Section \ref{secConclusion} concludes the paper.

\section{Model design}

\begin{figure*}
    \centerline{\includegraphics[width=\textwidth]{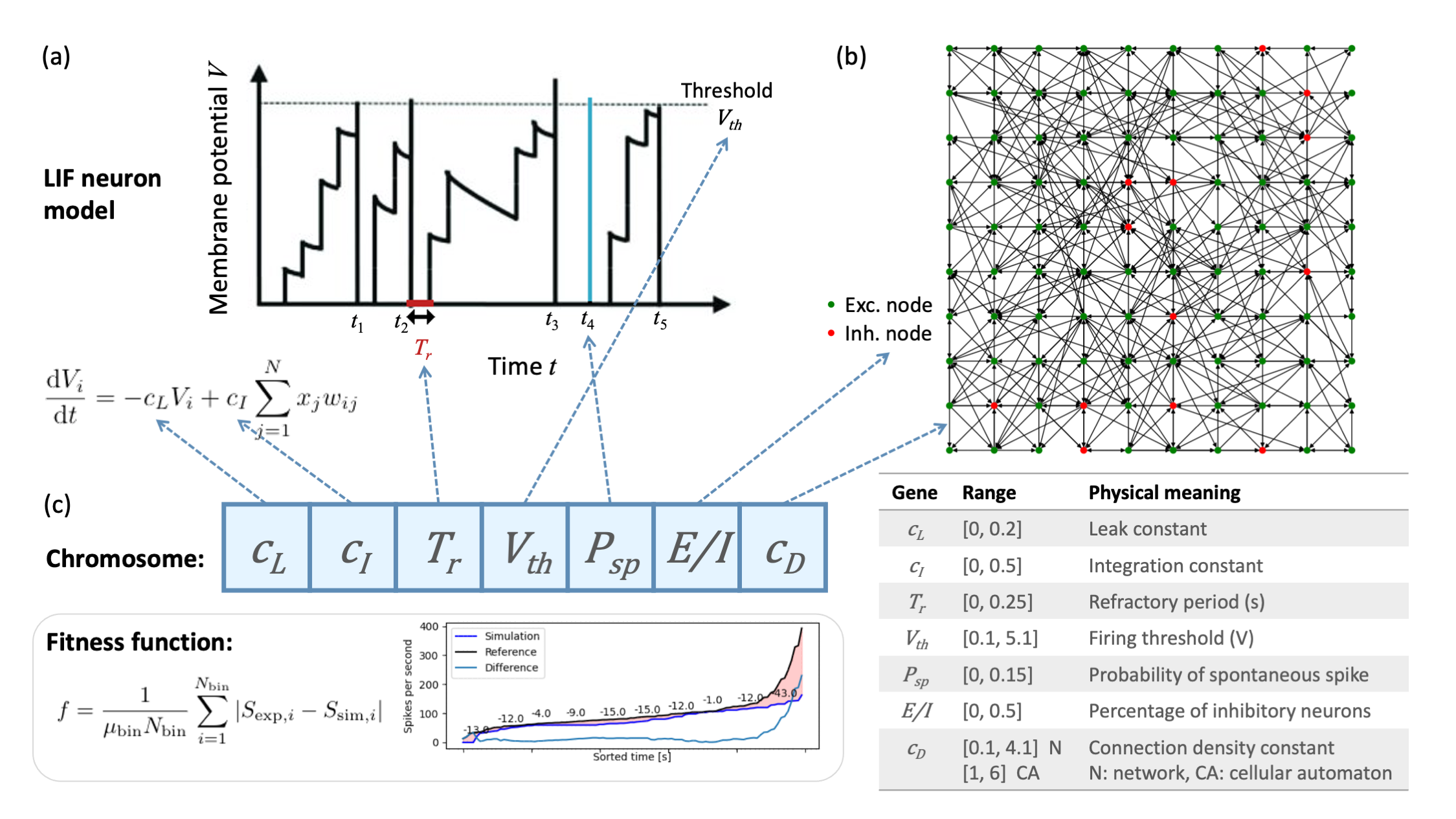}}
    \caption{Schematic of the model used in this study. (a) Response of the LIF model to a sequence of input spike trains. Each time the membrane potential reaches the threshold, the neuron spikes and its potential returns to the rest potential. (b) Schematic of a network model with $c_D=2.1$. (c) Chromosome depicting what parts of the model each gene describes, along with a list of the physical meaning and range of each gene, and the fitness function used in the EA.}
\label{figSchematic}
\end{figure*}

The models developed in this work consist of many simple spiking neuron models connected with different connectivity topologies\footnote{The code used in this study is publicly available online here: https://github.com/SocratesNFR/Evolving-spiking-neuron-cellular-automata-and-networks-to-emulate-in-vitro-neuronal-activity.}.
This section first presents the spiking neuron model, namely the leaky integrate-and-fire (LIF) neuron, and then describes how multiple neuron models were combined into larger systems, modeled as CAs and networks.

\subsection{Neuron model}

The neuron model used in this study is a simplified version of the leaky integrate-and-fire (LIF) neuron.
The LIF neuron combines three components of neuronal behavior into a simplified model: the integration of input signals, the threshold-based firing behavior, and the ``leaky'' membrane producing a slow decay in membrane potential.
A schematic of the response of the LIF model to a sequence of input spikes is shown in Fig.~\ref{figSchematic}(a), with each of the times along the horizontal axis representing spike times.
The membrane potential $V$ of the LIF neuron increases as input spikes arrive, and then slowly decreases back toward the resting potential with the characteristic resistor--capacitor (RC) discharging curve.
When the membrane potential reaches a threshold $T$, the neuron spikes, causing its output state to change from $x=0$ to $x=1$, and the membrane potential returns to the resting potential.

The LIF model is described by
\begin{equation}
    \tau_m \frac{\mathrm{d}V_i}{\mathrm{d}t} = c_L(E_L - V_i) + c_I\sum_{j=1}^N x_j w_{ij},
    \label{eqLIF}
\end{equation}
where $V_i$ is the membrane potential of neuron $i$;
$\tau_m$ is the membrane time constant, which describes the rate at which the membrane potential changes;
$t$ is time;
$c_L$ and $c_I$ are the leak and integration constants, which define the proportional contributions of the leaky current and the inputs, respectively, to the change in the membrane potential;
$E_L$ is the resting membrane potential;
$x_j$ is the output from neuron $j$;
and $w_{ij}$ is the weight of the connection between neurons $i$ and $j$.
The leak and integration constants were included in the model genome.

Here, this model was simplified by assuming $\tau_m = 1$ and $E_L = 0$, giving
\begin{equation}
    \frac{\mathrm{d}V_i}{\mathrm{d}t} = -c_LV_i + c_I\sum_{j=1}^N x_j w_{ij}.
    \label{eqSimpLIF}
\end{equation}
Additionally, a step activation function with a threshold of $V_i = T$ was used to convert the voltage $V_i$ of neuron $i$ to its state $x_i$ ($=0$ when not firing, $1$ when firing),
with the voltage was reset to the resting potential $V_i = 0$ when the threshold is reached.
After firing, the neuron enters a refractory period where it is unable to fire again.
The duration of the refractory period and the firing threshold were included in the genome.

To ensure ongoing activity in the network, the neurons were set to fire randomly with a certain probability.
Additionally, the neurons were divided into excitatory and inhibitory sub-populations, where outputs from excitatory neurons contribute to increasing the membrane potential and inhibitory to decreasing.
These two parameters---the probability of firing spontaneously and the proportion of inhibitory neurons---were included in the genome.
The overall behavior of the model is schematically shown in Fig.~\ref{figSchematic}, with the genome and fitness function shown in Fig.~\ref{figSchematic}(c).

\subsection{Neighborhoods and connectivity} \label{secConnectivity}

In this study, two types of system models were explored: a CA and a network model with connections formed based on geometric distance.
In both cases, the individual elements of the model---the cells in the CA and nodes in the network model---are all represented by the LIF model described above.
Thus, the difference between the two models is how these elements are connected.
The density of connections in each system is described by a density constant, which is among the evolved parameters and is defined differently for each system.

The cells of the CA have a Moore (eight-connected) neighborhood, and the density constant for the CA is defined as the neighborhood radius.
The network model topology was obtained by adding unweighted connections with a probability that decreases distance.
For the purposes of spatial representation in the network, the nodes are considered to be arranged in a two-dimensional grid.
For each pair of nodes, the likelihood of a connection between the nodes, in each direction, is given by
\begin{equation}
    P_{ij}(d_{ij}) = e^{-\left( \frac{d_{ij}}{c_D} \right)^2},
\end{equation}
where
$P_{ij}$ is the probability of a directed connection existing from node $i$ to node $j$,
$d_{ij}$ is the Euclidean distance between nodes $i$ and $j$, and
$c_D$ is the density constant.
The density constant was allowed to take on integer values ranging from 1 to 6 in the CA model and continuous values ranging from 0.1 to 4.1 in the network model.
In the network model, the same genome can produce different models because of the probability-based method of forming connections;
to avoid loss of the beneficial behavior of the top individuals, the connectivity was retained when elites were passed on to the next generation (see Sec.~\ref{secEvolution} for evolution details).

Both of these system models were generated with more nodes than recording channels in the experimental data.
The dimension of the models was set to $10 \times 10$ nodes or cells.
To evaluate the activity of the networks, 60 nodes were selected to roughly match the grid arrangement of the recording channels in the data.
In this way, the system models contained a proportion (41\%) of nodes whose activity was not observed, in a manner analogous to the subsampling that occurs by nature of the recording method used to obtain the biological neural data, though to a much greater extent.

\section{Emulation of neural behavior}

The goal of this study was to use activity recorded from neurons in vitro as target activity to be reproduced by our model systems.
Dissociated cultures of cortical neurons show complex network spiking activity, with different cultures exhibiting different activity modalities as they mature in vitro \cite{Gross1999,Heiney2021}.
The aim was not to precisely capture the dynamics of the biological neural networks with these models but to use a top-down approach with the aim of finding model parameters that yield similar spiking patterns in very simple network models.
This section will present an overview of the target data \cite{Wagenaar2006repertoire} used to drive the model evolution and the evolutionary process used to search the parameter space of CA and network models for models matching the target activity.

\subsection{In vitro neural data} \label{secData}

\begin{figure}
    \centerline{\includegraphics[width=3.5in]{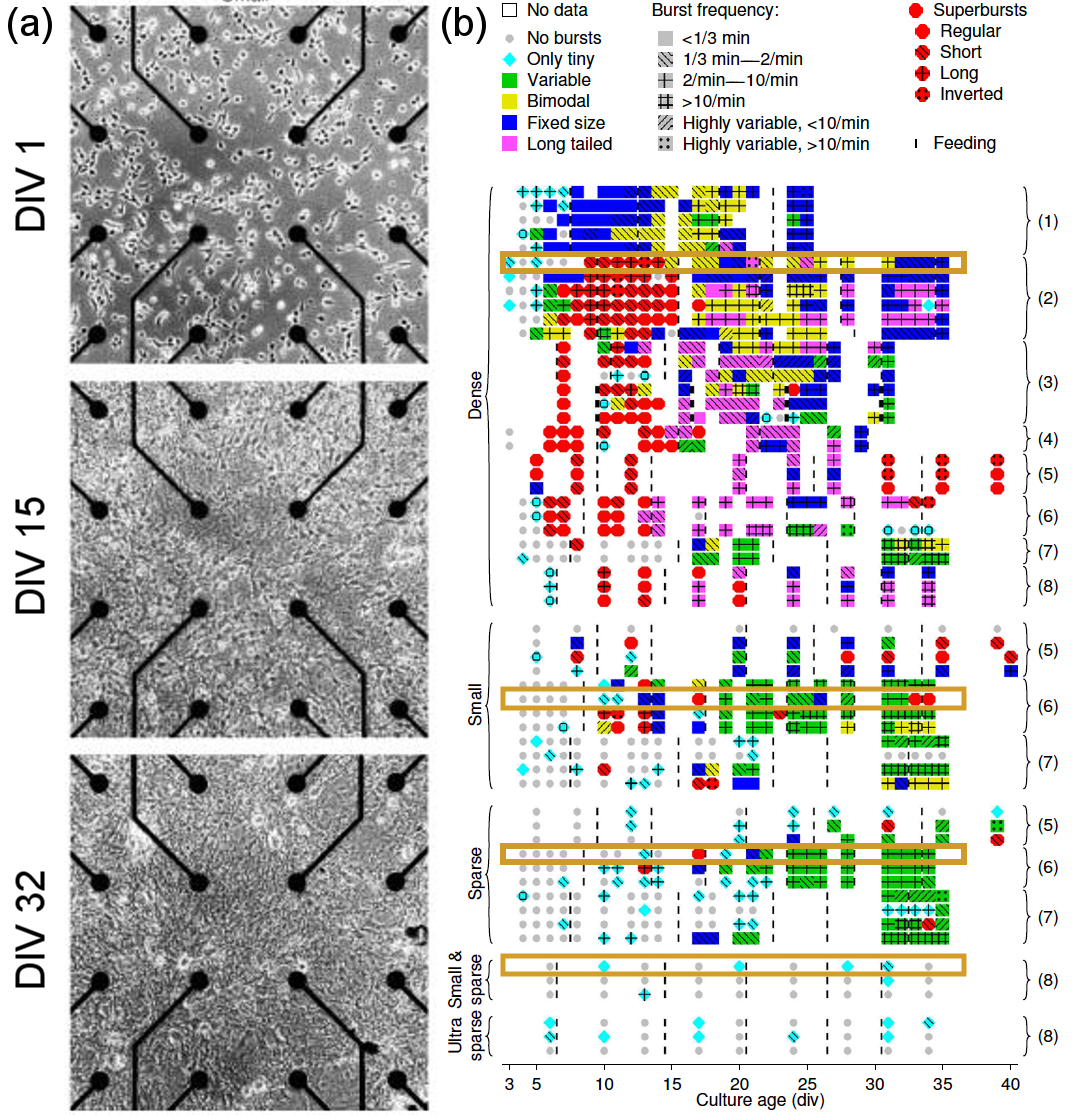}}
    \caption{Source of target data used in this study. (a) Phase contrast images of a Small cortical culture at DIVs 1, 15, and 32. (b) Classification of bursting patterns observed in networks of different densities. See Wagenaar et al.~\cite{Wagenaar2006repertoire} for an explanation of the different behaviors. The highlighted cultures (Dense culture 2-1, Small culture 6-2, Sparse culture 6-1, and Small \& sparse culture 8-1) were selected for use in our evolutionary process. Figure adapted from Wagenaar et al.~\cite{Wagenaar2006repertoire}.}
    \label{figWagenaar}
\end{figure}

The data used in this study was obtained from a comprehensive investigation by Wagenaar et al.~\cite{Wagenaar2006repertoire} into the varied spiking and bursting behaviors of in vitro cortical networks over a range of different plating densities and areas as they matured over five weeks in vitro.
A brief overview of their experiments and the observed behaviors will be presented here; details can be found in their paper \cite{Wagenaar2006repertoire}.
Dissociated cultures of cortical cells from rat embryos were prepared at five different combinations of seeding densities (ranging from 3,000 to 50,000 neurons plated on areas of 30 to 75 mm$^2$; referred to as Dense, Small, Sparse, Small \& sparse, and Ultra sparse).
Example phase contrast images of a Small culture at 1, 15, and 32 days in vitro (DIVs) are shown in Fig.~\ref{figWagenaar}(a); as shown in these images, the neurons first adhere to the substrate, then spontaneously begin to form an interconnected network, and the MEA channels record only a sample of the network activity.

The effect of plating density on the electrophysiological activity of the networks was observed as the cultures matured over the first five weeks of their development, with activity recorded for 30 min on most days using a 60-channel MEA.
Dissociated cortical cultures prepared in this way typically begin to show spontaneous spiking activity around 4--7 DIV,
and the onset of population bursting activity, where the network shows repetitive synchronized spikes across multiple electrodes, varies widely with culture density.
Fig.~\ref{figWagenaar}(b) shows the burst pattern classifications given by Wagenaar et al.~\cite{Wagenaar2006repertoire} to each culture on each day of recording.
As demonstrated here, the cultures showed a wide range of bursting behaviors across densities and over the course of their maturation.

For this study, we selected illustrative cultures of two densities (Small culture 6-2 and Small \& sparse culture 8-1) and evolved our models with target data from six time points during their maturation (DIVs 10, 13, 17, 24, 28, and 31).
These two densities showed very different behavior, with the Small cultures tending to display variable bursts from around DIV 20 and the Small \& sparse cultures showing no bursts or only tiny bursts.
To obtain a broader perspective on how density affects the evolutionary results, we also selected cultures from the Dense and Sparse classes and evaluated our model behavior for target data from two time points (DIVs 10 and 31, matching the first and last recording time points for the other two densities).

\subsection{Evolution of models to match spiking patterns}
\label{secEvolution}

To obtain models that show behavior comparable to the spiking behavior captured in the data, we used a generational EA with rank order selection, uniform crossover, and mild elitism, and the target parameters were encoded as floating-point genes.
As shown in Fig.~\ref{figSchematic}(c), the evolved parameters included in the genome were:
(1) the leak $c_L$ and (2) integration $c_I$ constants (see Eq.~\ref{eqSimpLIF}),
(3) the duration of the refractory period,
(4) the firing threshold,
(5) the probability a neuron spontaneously generates a spike,
(6) the proportion of inhibitory neurons in the population, and
(7) the density constant (see Section \ref{secConnectivity}).
The parameters selected for evolution were represented using genotype mapping with real-valued genes representing each parameter normalized to range from 0 to 1.
The upper and lower bounds of each gene are given in the table in Fig.~\ref{figSchematic}(c).
These bounds were obtained empirically by evaluating the behaviors that arose from extreme values of the parameters and selecting for eliminating certain behaviors.

The fitness of each model was evaluated by obtaining the ``average difference'' between sorted spike counts per bin obtained from the simulation and the experimental data.
The spike count curves used for this calculation were obtained as follows.
First, the data were divided into time bins of 1 s, and the spike counts were obtained for each bin.
The bins were then sorted in ascending order of spike count for each dataset.
Each experiment--simulation pair of sorted time bins was then compared by taking the difference, and this difference was averaged over all time bins.
An example of the how the average difference was computed is shown in the plot in Fig.~\ref{figSchematic}(c).
The goal of the EA was to minimize the average difference between the model data and the target experimental data, as described by
\begin{equation} \label{eqFitness}
    f = \frac{1}{\mu_{\mathrm{bin}} N_{\mathrm{bin}}} \sum_{i=1}^{N_{\mathrm{bin}}} |S_{\mathrm{exp},i} - S_{\mathrm{sim},i}|,
\end{equation}
where
$f$ is the objective function to be minimized;
$i$ is the time bin index;
$\mu_{\mathrm{bin}}$ is the spike count averaged over all time bins;
$N_{\mathrm{bin}}$ is the number of time bins in the considered data; and
$S_{\mathrm{exp},i}$ and $S_{\mathrm{sim},i}$ are the $i$th lowest spike counts in the binned experimental and simulation data, respectively.
In early tests of the model, we implemented a fitness function that had a similar spatial component of the ``average distance'' across nodes, where the spike counts were considered on each recording channel or node; however, the models were unable to achieve the desired spatial heterogeneity, and so this fitness function was abandoned (see Section~\ref{secLimitations}).

An initial population of $N=60$ models was generated and evaluated using the fitness function.
The fitness of each of these models was evaluated by running the model for $T=60$~s, where one time step in the simulation was considered to correspond to 40~ms in the experimental data, and computing its fitness with Eq.~\ref{eqFitness}.
The top-performing 50\% of models were retained as the parent generation, and crossover and mutation were performed to produce the next generation.
Crossover was performed by randomly pairing the selected top individuals and randomly selecting one of the two parents to contribute each gene.
This process was repeated until the population reached the same size $N$ as the initial population.
After crossover, mutation was performed by assigning a randomly generated value to a gene with a mutation probability of 10\%.
Additionally, the top-performing 5\% of all individuals in each generation is passed on without any crossover or mutation to retain any particularly high-performing individuals (elitism). The evolution process was iterated for 80 generations, and the top-performing model was selected for more detailed analysis at the end of the process.
For each combination of model (CA, network), culture density (Dense, Small, Sparse, Small \& sparse), and recording DIV, 10 trials were performed.

\section{Results and discussion}
\label{secResults}

This section presents the results of our experiments.
The EA was able to produce models with high fitness when the patterns of spiking activity were of fairly low complexity, and the fitness decreased with increasing complexity (higher culture density and later maturation time points).
However, we were able to capture network-wide bursting events in our models despite their simple topology and homogeneity.
An assessment of the genomes of the top-performing models reveals the mechanisms by which our models emulate the in vitro behavior, with greater levels of activity and more complex patterns correlating with higher density constants, lower refractory periods, and increased excitability.


\subsection{Model behavior over generations}

\begin{figure}
    \centerline{\includegraphics[width=3.5in]{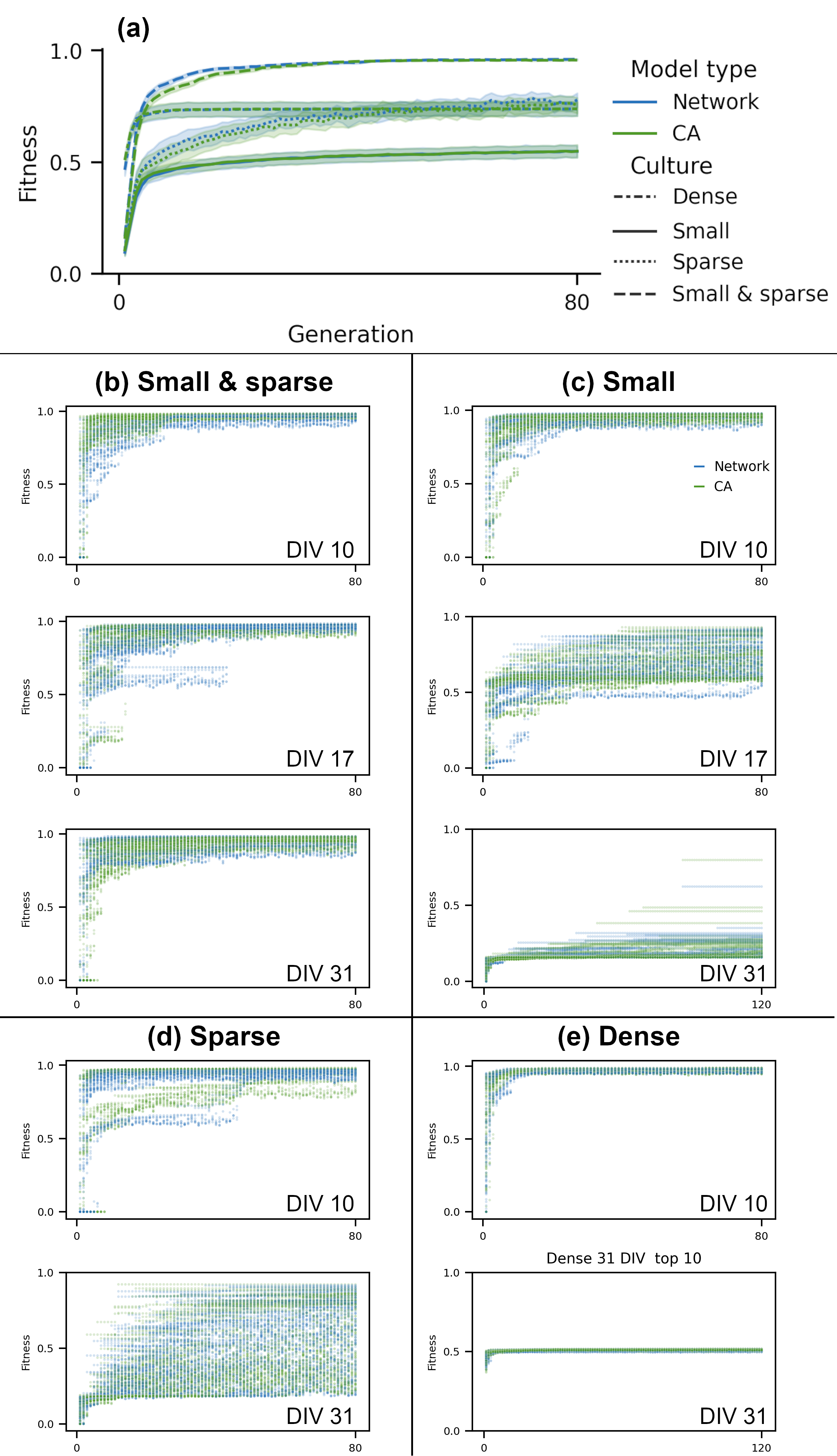}}
    \caption{Fitness over generations. (a) Fitness of the top 10 individuals in each trial for each combination of model and culture density. Lines represent the mean fitness of the top 10 individuals over all 10 trials and all considered recording time points, and the shaded areas represent one standard deviation above and below the mean.
    (b)--(e) Fitness scatter plots of the top 10 individuals in every trial over successive generations at select DIVs for each culture density:
    (b) Small \& sparse,
    (c) Small,
    (d) Sparse,
    (e) Dense.
    Note that the DIV 31 trials for the Small and Dense cultures have more generations because of the greater amount of activity produced by these cultures.}
    \label{figFitnessVsGen}
\end{figure}

Fig.~\ref{figFitnessVsGen}(a) shows how the fitness of the different models changed over the course of the evolution.
As shown in this figure, the network and CA models were able to achieve a similar overall fitness, with the fitness reaching peaks of approximately 0.9, 0.7, 0.5, and 0.7 for the Small \& sparse, Sparse, Small, and Dense cultures, respectively.
In the case of the Small \& sparse culture, the fitness trend plateaued near its peak value around generation 30, with the CA models reaching this plateau slightly earlier than the network models.
The Small models and Sparse models showed a sharp rise in the first five generations and a gradual increase over the remaining generations, with the network models slightly outperforming the CA models in the case of the Sparse models.

The peak fitness tended to decrease with increasing culture density;
this is not surprising, as the higher-density cultures tend to show more complicated and varied patterns of activity.
As indicated in Fig.~\ref{figWagenaar}(b), the Small culture and the Sparse culture exhibited a range of bursting behaviors, with tiny bursts starting at DIV 10--13 and even showing some superbursts at certain points, 
whereas the Small \& sparse culture only occasionally fluctuated between tiny bursts and no bursts.
The more regular behavior shown by the lower-density cultures is easier to capture in our model system, and our fitness function can still give good performance even if bursts are not consistently produced by the models.

It is interesting to note that the CA and network models showed similar fitness trends, as it is well-known that topology plays a large role in determining the behavior of a dynamical system.
These results indicate that it is possible, with the right parameter tuning, to achieve interesting behaviors in a regularly connected neuron model despite the limitations imposed by such a model.
However, it is also worth noting that the network models tend to slightly outperform the CA models overall and perhaps would be able to achieve richer dynamics given more evolution time.

The results here indicate that the evolutionary process tends to find high-fitness models rather early on.
The fitness of all individuals in specific trials were investigated in more detail to evaluate the consistency of this trend.
Fig.~\ref{figFitnessVsGen}(b)--(e) shows scatter plots of the fitness of the top 10 individuals from all trials for a given culture density at the two time points considered for all four of the culture densities (DIVs 10 and 31), along with a third time point for the two cultures with more detailed analysis (DIV 17).
At DIV 10, the models tended to converge toward high-fitness solutions very quickly, with the more complex activity at later time points generally driving down the fitness score.
The following section will present the different spiking behaviors of a selection of the top-performing models.

\subsection{Similarity of model outputs to target spiking patterns}

\begin{figure}
    \centerline{\includegraphics[width=3.5in]{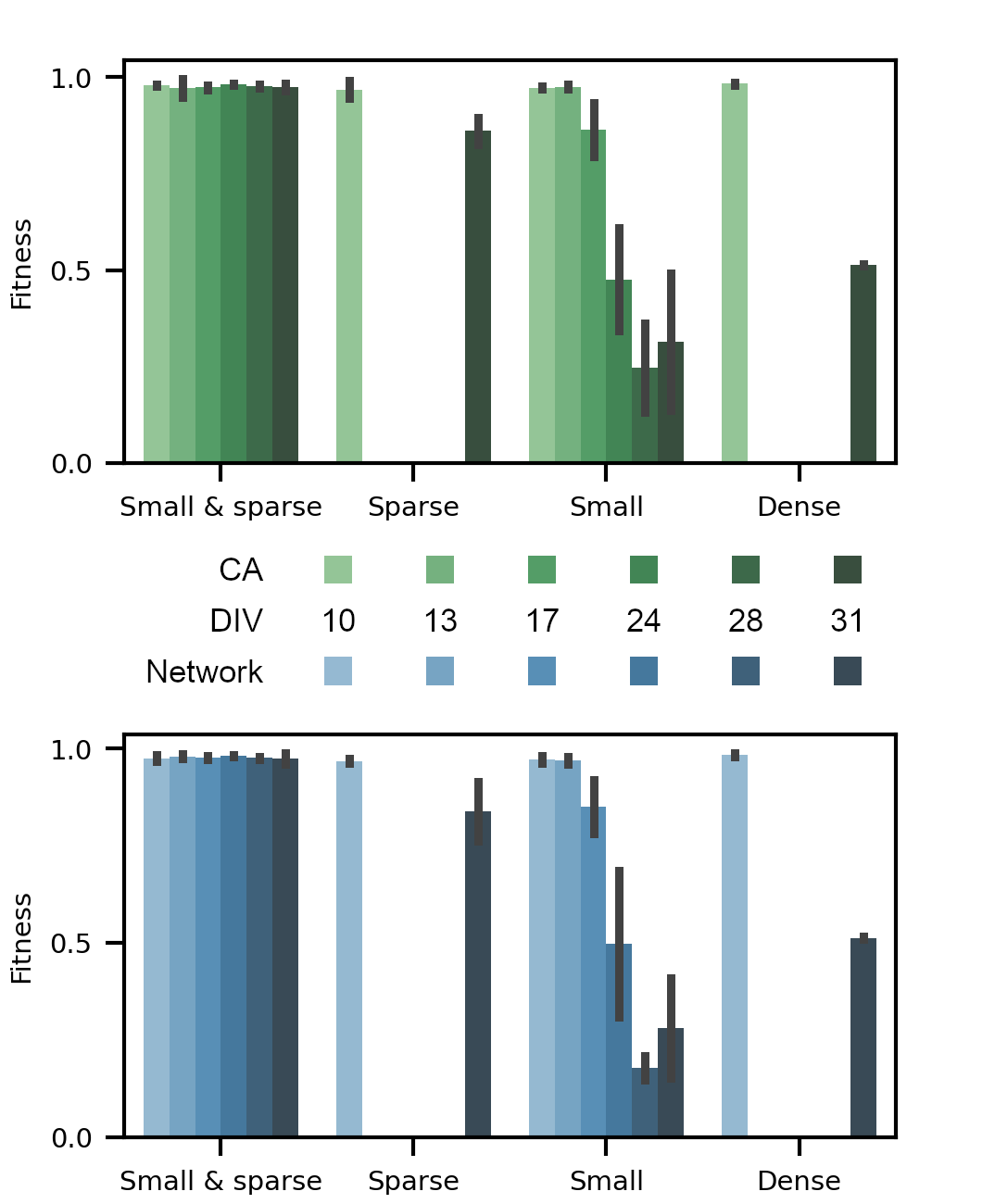}}
    \caption{Fitness of the top 10 individuals of each trial at the considered recording time points for each culture density: (a) CA models, (b) network models. Plotted values represent the mean fitness of the top 10 individuals over all 10 trials, and the error bars represent one standard deviation above and below the mean.}
    \label{figFitnessVsDIV}
\end{figure}

Fig.~\ref{figFitnessVsDIV} shows the mean fitness of the 10 highest-performing models in each trial over the six considered recording time points for each culture density.
The Small \& sparse models showed consistent fitness scores of approximately 0.9 across all DIVs.
In contrast, the Small models showed high fitness scores of approximately 0.9 at DIVs 10 and 13, with the fitness generally decreasing rapidly with progressive DIVs beyond DIV 17.
The Sparse and Dense models also showed a drop in fitness between the early and late recording time points, with the Dense models showing a sharper decrease in fitness.
As discussed in the previous section, the CA and network models showed similar fitness scores.

The lower performance of the models corresponds to changes in the bursting behavior of the cultures.
In the three higher-density cultures, the low scores correspond to time points where the culture showed variable or bimodal bursts, and the higher scores correspond to cases of tiny and fixed-size bursts.
These results provide additional evidence that our models achieve good performance only up to a certain level of spiking complexity.

\begin{figure}
    \centerline{\includegraphics[width=3.5in]{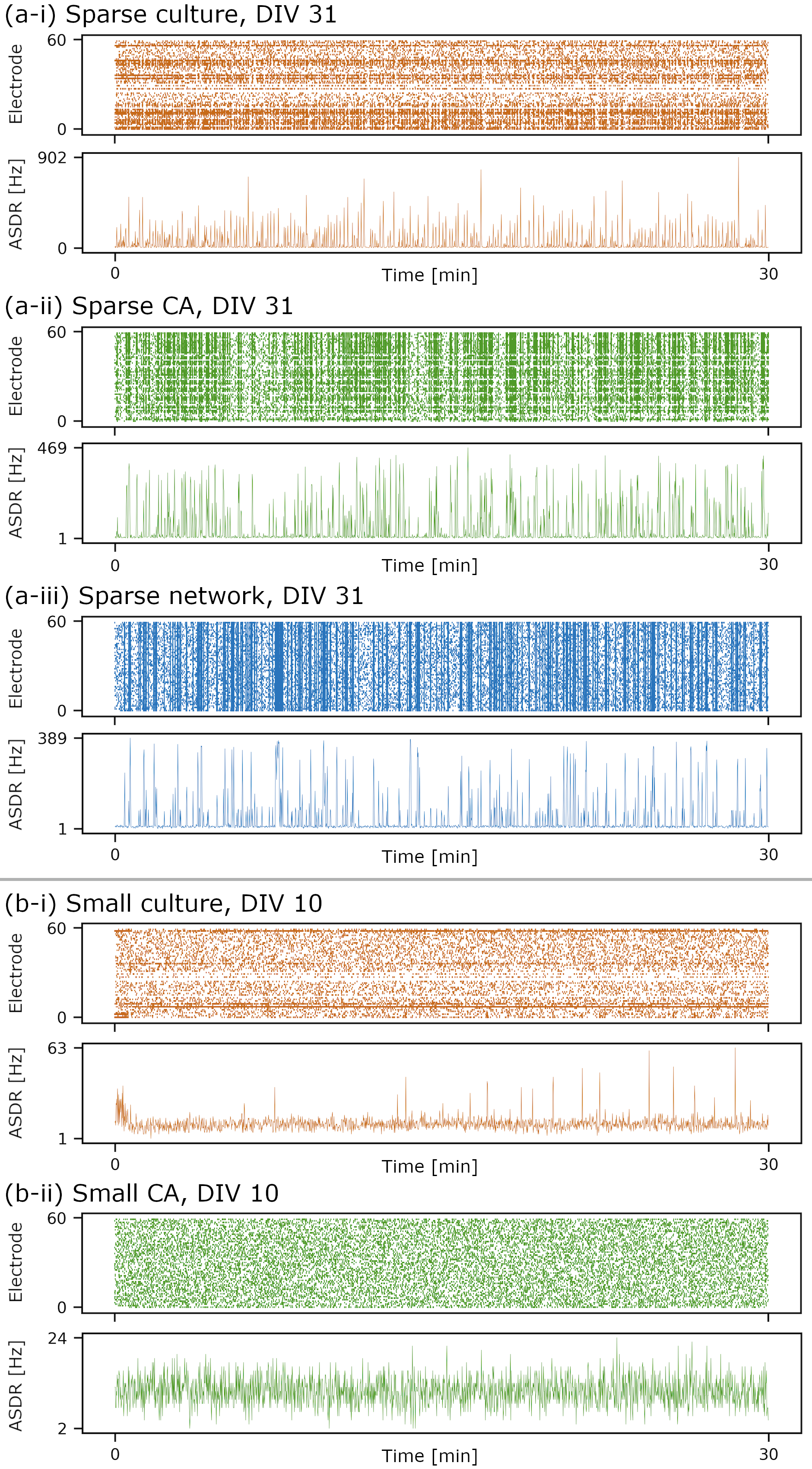}}
    \caption{Example spiking behavior of top individuals on DIV 31. (a-i) Sparse culture activity on DIV 31, along with the (a-ii) CA and (a-iii) network obtained with this target data. (b-i) Small culture activity on DIV 10, along with the activity for the (b-ii) CA obtained with this target data. ASDR: Array-wide spike detection rate (see \cite{Wagenaar2006repertoire}).}
    \label{figRasters}
\end{figure}

However, it is important to note that our models were not incapable of demonstrating network-wide bursting behavior.
Fig.~\ref{figRasters} shows a few example raster plots for top-performing models run for 30 min, along with the corresponding array-wide spike detection rate (ASDR), which was used by Wagenaar et al.~\cite{Wagenaar2006repertoire} to represent bursting patterns in their data.
This measure is simply the number of spikes in the entire network per second.
As shown in Fig.~\ref{figRasters}(a-ii) and (a-iii), both the CA and network models were able to produce network-wide bursts, and they showed complex patterns of activity in these example cases.
The network model in this case showed more spatially varied activity, whereas the CA tended to be more spatially homogeneous.
Although the distribution of spiking across nodes did not always match that in the target data, the population-level activity often showed similar signatures.
There were also a number of cases where the system activity was not very complex, such as the case shown in Fig.~\ref{figRasters}(b).
In this case, there was a consistently low level of activity that did not show any noticeable spatiotemporal patterns, and this was observed for both the CA (Fig.~\ref{figRasters}(b-ii)) and network model (not shown).
In this particular case, the level of network-wide activity was quite low in the data, so the resulting models were able to achieve high fitness with rather temporally homogeneous behavior.

\subsection{Model genome}

\begin{figure*}
    \centerline{\includegraphics[width=0.9\textwidth]{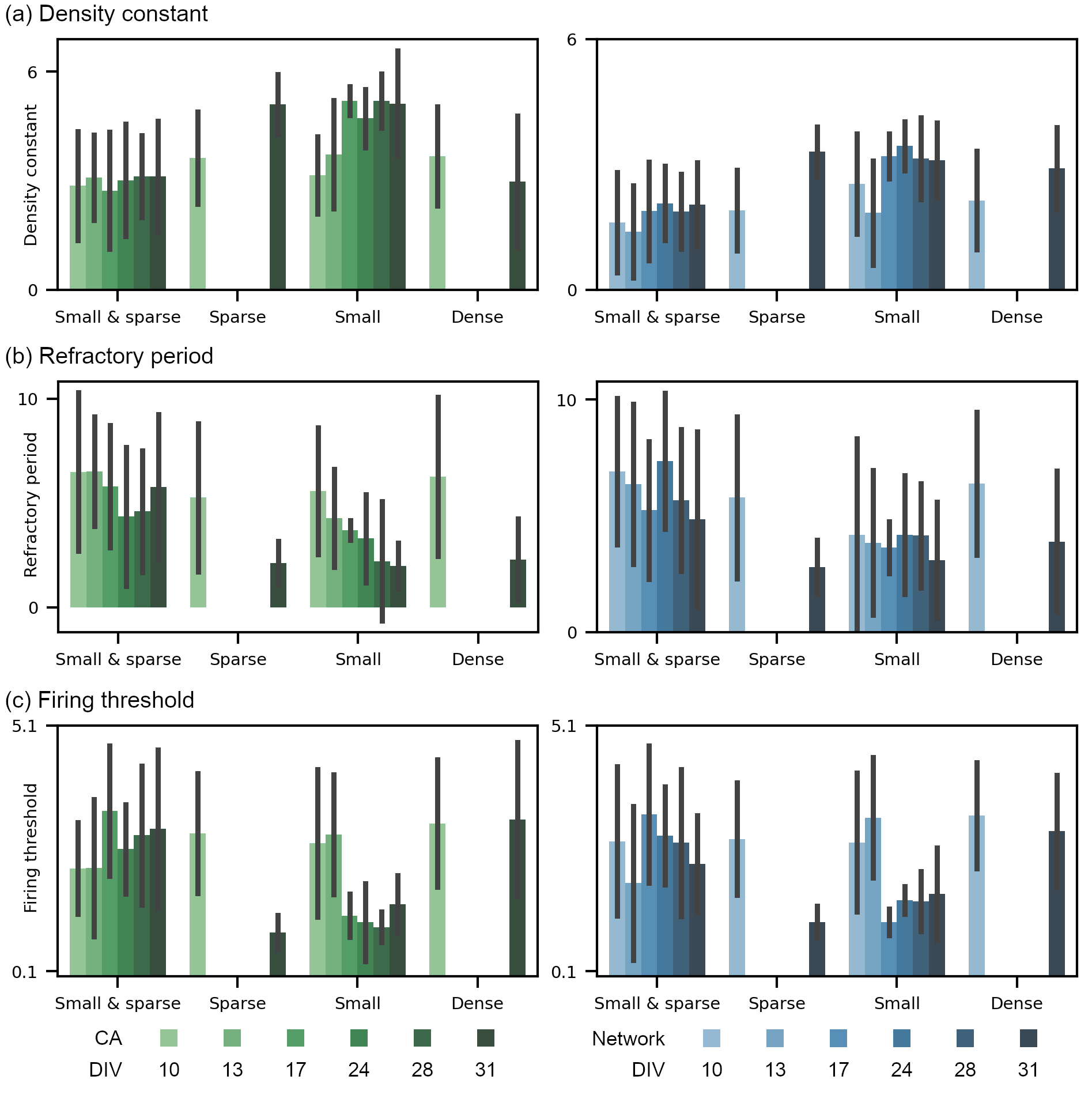}}
    \caption{Select genes averaged over the 10 highest-performing models of each trial, plotted for each culture density over the selected DIVs. Error bars represent standard deiviations. (a) Density constant. (b) Refractory period. (c) Firing threshold. The left- and right-hand columns show CA and network model results, respectively.}
    \label{figGenome}
\end{figure*}

The model genome of the top-performing individuals was evaluated to assess if there were any trends in the gene values.
A selection of genes is shown in Fig.~\ref{figGenome}.
Each plot shows the gene values averaged over the 10 highest-performing individuals in every trial, with error bars representing the standard deviation.
Gene values are plotted for each culture density over the selected DIVs.
The left column, in green, shows the CA results, and the right, in blue, shows the network results.

Most genes showed fairly large variability in their values, even when fitness trends were rather consistent across multiple trials of the same culture at the same time point.
Some interesting trends emerged in the selected genes: the density constant, refractory period, and firing threshold (Fig.~\ref{figGenome}(a)--(c), respectively).
The density constant tended to increase with increasing DIV (except for the Small \& sparse culture) and with increasing culture density, particularly in the case of the CA models.
This indicates that denser connections are needed to reproduce the more complex patterns of activity seen with increasing culture density and maturity.

The refractory period showed a great deal of deviation across cultures and recording time points, and tended to decrease over progressive DIVs.
Allowing neurons to fire more rapidly after a given spike allows a greater spike count and higher activity levels, which is consistent with the type of behavior that occurs in the in vitro cultures at later DIVs.

The firing threshold also showed a decreasing trend over progressive DIVs, though with a weaker correlation.
The trend in the ratio of inhibitory cells (results not shown) very closely followed that of the firing threshold, particularly with the sharp drop for the Small culture at DIV 17.
Reductions in both of these factors increase the excitability of the network, also allowing greater levels of activity and greater responsiveness to activity of presynaptic neurons.

The spontaneous firing probability remained roughly constant at very low values regardless of culture density, DIV, or model type, except in the case of the Dense culture at DIV 31.
In the case of the Dense culture at DIV 31, however, the spontaneous firing probability was rather high on average, near 5\%;
this indicates that the Dense models heavily relied on randomly generated spontaneous activity to approach the number of spikes seen in the Dense cultures.



\subsection{Limitations}
\label{secLimitations}

One of the main limitations of our approach is the overall homogeneity of the developed models.
In our models, each LIF neuron is described by the same parameters, and the neurons have the same or similar numbers of connections with other neurons.
In contrast, in the target in vitro biological system, the neurons do not show homogeneous properties or patterns of connectivity.
In particular, there is evidence that networks of biological neurons tend to show certain patterns of connectivity, such as small-worldness, that are understood to support efficient computation \cite{Heiney2021,Laughlin2003}.
Additionally, the fitness function used in this study considered only the temporal distribution of the spiking activity of the entire population, giving no preference to more spatially heterogeneous activity.

To evaluate whether a simple change to the fitness function would drive the models toward greater spatial heterogeneity, the fitness function was redesigned to include a spatial component as well as a temporal component.
The spatial component considered the average distance across the sorted mean firing rate per channel in the same manner as the original fitness function.
The results indicate that this simple change is not sufficient to drive the models to produce spatially heterogeneous activity, and the resulting models ultimately performed worse on the temporal component of the fitness function as well.

Thus, to achieve spatially variable patterns of activity, greater heterogeneity must be introduced into the models themselves, both in terms of varying node behavior across the network and varying the connectivity patterns in the system as a whole.
This can be done by sampling gene values from distributions to describe the parameters of the single-neuron models as well as to define the degree distribution of the network.
For example, one gene could define the type of distribution (e.g., Gaussian, Poisson, exponential) for a model parameter to take, and other genes could define the parameters for that distribution (e.g., mean, standard deviation).
Then each node in the model would have a different gene value drawn from these distributions.
These approaches could then be coupled with a fitness function that rewards spatial heterogeneity as well.

Another factor that would greatly expand the repertoire of behaviors achievable by our models would be to include axonal delays in our models.
Axonal delays are a major contributing factor to the information processing capabilities in neuronal networks, both biological \cite{Debanne2004} and artificial \cite{Izhikevich2006}.
Introducing heterogeneous delays in the model would increase the nonlinearity of the system and likely improve its performance as a reservoir.

The approach used in the current study is also limited by the amount of recording time selected for comparison in the fitness function.
Only 1 min was used of the full 30-min recordings for the sake of computation time.
However, this small observation window may have excluded some interesting patterns of behavior.

Finally, although the proposed fitness function focused on the temporal aspect of the spiking behavior, it did so in a way that may have sacrificed the relevant timescales and temporal relations in the data.
That is, the number of spikes per second was considered in the fitness function, but this may aggregate finer patterns of activity on the millisecond scale, and the time bin sorting may eliminate important temporal relations between initially consecutive bins.
The sorting was done to avoid overfitting, as we were interested in capturing the overall patterns of activity without precisely reproducing the observed data; however, it is likely that considering multiple scales of the spiking distribution over time (i.e., different size time bins) may aid in retaining information spanning timescales and temporal relations in the activity.

\section{Conclusion}
\label{secConclusion}

In this study, we applied an EA to emulate the spiking activity of in vitro neural cultures of different plating densities and at different time points during their maturation, with the aim of ultimately using models produced in this manner as computational substrates.
Our networks were able to produce activity similar to that of the cultures early in their maturation or cultures with lower densities---under these conditions, the neurons show fewer and smaller bursts.
Although our networks were unable to produce the more varied and complex patterns of activity that arose in mature higher-density cultures, they were still able to produce short network-wide bursts, demonstrating that even relatively simple and homogeneous networks can be tuned to produce somewhat complex activity.

This evolutionary approach will lay the foundation for future work on developing unconventional computing systems that are able to emulate targeted biological behaviors considered beneficial for computation.
Future work will involve first advancing the present methods, by, for example, exploring the use of heterogeneous system models and different fitness functions, and then evaluating the performance of the output models on computational tasks.

\section*{Acknowledgment}

The authors thank the Nordic Center for Sustainable and Trustworthy AI Research (NordSTAR), as well as the SOCRATES and DeepCA projects, for their support in this project. The authors are also grateful for the support of the OsloMet AI Lab.

\bibliographystyle{IEEEtran}
\bibliography{refs}

\end{document}